\documentclass[10pt, a4paper]{article}
\usepackage{lrec}
\usepackage{graphicx}
\usepackage{tabularx}
\usepackage{soul}

\usepackage{amsmath}
\usepackage{epstopdf}
\usepackage[latin1]{inputenc}

\usepackage{hyperref}
\usepackage{xstring}

\newcommand{\comment}[1]{}

\title{Localization of Fake News Detection via Multitask Transfer Learning}

\name{Jan Christian Blaise Cruz, Julianne Agatha Tan, and Charibeth Cheng}

\address{Center for Language Technologies (CeLT), De La Salle University, Manila \\
         2401 Taft Ave, Malate, Manila, Philippines \\
         \{jan\_christian\_cruz, julianne\_tan, charibeth.cheng\}@dlsu.edu.ph\\}

\abstract{
    The use of the internet as a fast medium of spreading fake news reinforces the need for computational tools that combat it.
	Techniques that train fake news classifiers exist, but they all assume an abundance of resources including large labeled datasets and expert-curated corpora, which low-resource languages may not have. 
	In this work, we make two main contributions:
	First, we alleviate resource scarcity by constructing the first expertly-curated benchmark dataset for fake news detection in Filipino, which we call ``Fake News Filipino.''
    Second, we benchmark Transfer Learning (TL) techniques and show that they can be used to train robust fake news classifiers from little data, achieving 91\% accuracy on our fake news dataset, reducing the error by 14\% compared to established few-shot baselines.
    Furthermore, lifting ideas from multitask learning,  we show that augmenting transformer-based transfer techniques with auxiliary language modeling losses improves their performance by adapting to writing style.
    Using this, we improve TL performance by 4-6\%, achieving an accuracy of 96\% on our best model.
    Lastly, we show that our method generalizes well to different types of news articles, including political news, entertainment news, and opinion articles.  \\ \newline \Keywords{Statistical and Machine Learning Methods, Language Modelling, Document Classification and Text Categorization} }

\begin{document}

\maketitleabstract

\section{Introduction}
There is a growing interest in research revolving around automated fake news detection and fact checking as its need increases due to the dangerous speed fake news spreads on social media \cite{perezAutofake2018}. With as much as 68\% of adults in the United States regularly consuming news on social media\footnote{https://www.journalism.org/2018/09/10/news-use-across-social-media-platforms-2018/}, being able to distinguish fake from non-fake is a pressing need. 

Numerous recent studies have tackled fake news detection with various techniques. The work of \newcite{stance2017bourgonje} identifies and verifies the stance of a headline with respect to its content as a first step in identifying potential fake news, achieving an accuracy of 89.59\% on a publicly available article stance dataset. The work of \newcite{multi2018karimi} uses a deep learning approach and integrates multiple sources to assign a degree of ``fakeness'' to an article, beating representative baselines on a publicly-available fake news dataset.

More recent approaches also incorporate newer, novel methods to aid in detection. The work of \newcite{confortiStance2018} handles fake news detection as a specific case of \textit{cross-level stance detection}. In addition, their work also uses the presence of an ``inverted pyramid'' structure as an indicator of real news, using a neural network to encode a given article's structure.

While these approaches are valid and robust, most, if not all, modern fake news detection techniques assume the existence of large, expertly-annotated corpora to train models from scratch. Both \newcite{stance2017bourgonje} and \newcite{confortiStance2018} use the Fake News Challenge\footnote{http://www.fakenewschallenge.org/} dataset, with 49,972 labeled stances for each headline-body pairs. \newcite{multi2018karimi}, on the other hand, uses the LIAR dataset \cite{liar2017wang}, which contains 12,836 labeled short statements as well as sources to support the labels.

This requirement for large datasets to effectively train fake news detection models from scratch makes it difficult to adapt these techniques into low-resource languages. Our work focuses on the use of Transfer Learning (TL) to evade this data scarcity problem.

We make two main contributions:

We construct the first fake news dataset in the low-resourced Filipino language, alleviating data scarcity for research in this domain. We call this dataset ``Fake News Filipino.'' 
    
We benchmark TL techniques such as ULMFiT \cite{ulmfit2018howard}, BERT \cite{bert2018devlin}, and GPT-2 \cite{gpt2018radford,gpt22019radford} and show that they perform better compared to few-shot techniques by a considerable margin. 

Furthermore, we show that auxiliary language modeling losses \cite{chronopoulou2019embarrassingly,liu2019multi} allows transformers to adapt to the writing style of downstream tasks, which produces more robust fake news classifiers. 

\section{Fake News Dataset}

\begin{table*}[t!]
\centering
\begin{tabular}{llllll}
\hline
  Split & Label & Documents & Tokens & Unique Tokens & OOV Tokens \\ \hline
  Train & - & 2,244 & 468,056 & 41,570 &  \\
  Test  & - & 962   & 200,472 & 24,978 & 8,807 (17.48\%) \\
  \hline
  Train + Test & Real & 1,603 & 447,401 & 35,959 &  \\
  & Fake & 1,603 & 221,127 & 27,371 & 14,418 (28.62\%) \\
  \hline
  Train & Real & 1,121 & 312,047 & 29,619 &  \\
  & Fake & 1,123 & 156,009 & 22,385 & 11,951 (28.75\%) \\
  \hline
  
\end{tabular}
\caption{\label{fnf} Statistics for the Fake News Filipino dataset. The top division of the table describes corpus statistics for train and test splits. The middle describes token statistics for real and fake articles in the entire dataset. The bottom describes token statistics for real and  fake articles in the training split of the dataset. Training set size is 70\% of the total size of the dataset. To produce the training set, the full dataset is loaded into Pandas and is shuffled in place with a random seed of 42.}
\end{table*}
We remedy the lack of a proper, curated benchmark dataset for fake news detection in Filipino by constructing and producing what we call ``Fake News Filipino.'' The dataset is composed of 3,206 news articles, each labeled \textit{real} or \textit{fake}, with a perfect 50/50 split between 1,603 real and 1,603 fake articles, respectively.

Fake news articles were sourced from online sites that were tagged as \textit{fake news sites} by the non-profit independent media fact-checking organization Verafiles\footnote{https://verafiles.org/} and the National Union of Journalists in the Philippines\footnote{https://nujp.org/} (NUJP). Real news articles were sourced from mainstream news websites in the Philippines, including Pilipino Star Ngayon\footnote{https://www.philstar.com/pilipino-star-ngayon}, Abante\footnote{https://www.abante.com.ph/}, and Bandera\footnote{https://bandera.inquirer.net/}.

The dataset is primarily in Filipino, with the addition of some English words commonly used in Filipino vernacular. Token and corpus statistics for the dataset can be found on Table \ref{fnf}.

We construct the dataset by scraping our source websites, encoding all characters into UTF-8. Preprocessing was light to keep information intact: we retain capitalization and punctuation, and do not correct any misspelled words. We do not pre-tokenize the dataset, but in order to compute the dataset statistics in Table \ref{fnf}, we use the Moses Tokenizer \cite{koehn2007moses}.

\begin{table}[t!]
\centering
\begin{tabular}{lll|l}
  \hline
   & Real & Fake & Unique Fake  \\ 
  \hline
  1 & si      & si        & koponan \\
  2 & naman   & naman     & torneo \\
  3 & kay     & kay       & PVF \\
  4 & daw     & daw       & Volleyball \\
  5 & lang    & Duterte   & barahang \\
  6 & Duterte & lang      & panalo-talo \\
  7 & noong   & ayon      & rebounds \\
  8 & ayon    & umano     & Mayweather \\
  9 & I       & source    & asosasyon \\
  10 & rin    & video     & Coloma \\
  11 & umano  & ilang     & Bimby \\
  12 & nang   & Miss      & Romasanta \\
  13 & ilang  & po        & MVP \\
  14 & source & pangulong & open \\
  15 & yung   & rin       & kings \\
  \hline
  
\end{tabular}
\caption{\label{fnfwordsfreq} Top words in the Fake News Filipino dataset by frequency. Real and fake news articles have very similar top words, sharing 10 words with the highest frequency. The right hand shows the top fifteen words that only show in fake news articles (in the context of the dataset) by frequency.}
\end{table}

We also perform initial analyses on the FNF dataset. It can be seen that real news and fake news share very similar vocabularies, with 10 out of the top 15 words by frequency shared by the two classes. Looking at the entirety of the real and fake vocabularies, a large overlap is also seen. In the full dataset, only 28.62\% of words appear solely in fake news articles do not appear in real news article. In the training split, only 28.75\% of words unique to fake articles do not appear in real news. As the dataset only contains 1,603 real and fake articles respectively, this overlap is expected to shrink as more data is collected. This tells us that the presence of certain words do not guarantee that an article will be classified as real or fake. 

We also list the top 15 words by frequency that only appear in fake news articles. We find that 5 out of 15 are proper names, with the rest of the 10 being neutral words. This further supports the idea that word choice does not immediately point towards "fakeness" or "realness" of an article.

The list of top words by frequency, along with the top 15 words by frequency that only appear in fake news articles, is shown in Table \ref{fnfwordsfreq}.

\section{Methods}
We provide a baseline model as a comparison point, using a few-shot learning-based technique to benchmark transfer learning against methods designed with low resource settings in mind. After which, we show three TL techniques that we adapted to the task of fake news detection.

\subsection{Baseline}
We use a siamese neural network, shown to perform state-of-the-art few-shot learning \cite{kochSiamese2015}, as our baseline model.

A siamese network is composed of weight-tied twin networks that accept distinct inputs, joined by an energy function, which computes a distance metric between the representations given by both twins. The network could then be trained to differentiate between classes in order to perform classification \cite{kochSiamese2015}.

We modify the original to account for sequential data, with each twin composed of an embedding layer, a Long-Short Term Memory (LSTM) \cite{hochreiterLSTM1997} layer, and a feed-forward layer with Rectified Linear Unit (ReLU) activations. 

Each twin embeds and computes representations for a pair of sequences, with the prediction vector $p$ computed as:
\begin{equation}
	\textnormal{p} = \sigma(W_{\textnormal{out}}|o_1 - o_2| + b_{\textnormal{out}})
\end{equation}
where $o_i$ denotes the output representation of each siamese twin $i$ , $W_{\textnormal{out}}$ and $b_{\textnormal{out}}$ denote the weight matrix and bias of the output layer, and $\sigma$ denotes the sigmoid activation function. 

\subsection{ULMFiT}

ULMFiT \cite{ulmfit2018howard} was introduced as a TL method for Natural Language Processing (NLP) that works akin to ImageNet \cite{imagenet2015russakovsky} pretraining in Computer Vision. 

It uses an AWD-LSTM \cite{awdlstm2017merity} pretrained on a language modeling objective as a base model, which is then finetuned to a downstream task in two steps. 

First, the language model is finetuned to the text of the target task to adapt to the task syntactically. Second, a classification layer is appended to the model and is finetuned to the classification task conservatively. During finetuning, multiple different techniques are introduced to prevent catastrophic forgetting.

ULMFiT delivers state-of-the-art performance for text classification, and is notable for being able to set comparable scores with as little as 1000 samples of data, making it attractive for use in low-resource settings \cite{ulmfit2018howard}.

\subsection{BERT}

BERT is a Transformer-based \cite{transformer2017vaswani} language model designed to pretrain ``deep bidirectional representations'' that can be finetuned to different tasks, with state-of-the-art results achieved in multiple language understanding benchmarks \cite{bert2018devlin}. 

As with all Transformers, it draws power from a mechanism called ``Attention'' \cite{attention2015luong}, which allows the model to compute weighted importance for each token in a sequence, effectively pinpointing context reference \cite{transformer2017vaswani}. Attention allows the Transformer to refer to multiple positions in a sequence for context at any given time regardless of distance, which is an advantage over Recurrent Neural Networks (RNN).

BERT's advantage over ULMFiT is its bidirectionality, leveraging both left and right context using a pretraining method called ``Masked Language Modeling.'' In addition, BERT also benefits from being \textit{deep}, allowing it to capture more context and information. BERT-Base, the smallest BERT model, has 12 layers (768 units in each hidden layer) and 12 attention heads for a total of 110M parameters. Its larger sibling, BERT-Large, has 24 layers (1024 units in each hidden layer) and 16 attention heads for a total of 340M parameters.

\subsection{GPT-2}
The GPT-2 \cite{gpt22019radford} technique builds up from the original GPT \cite{gpt2018radford}. Its main contribution is the way it is trained. With an improved architecture, it learns to do multiple tasks by just training on vanilla language modeling. 

Architecture-wise, it is a Transformer-based model similar to BERT, with a few differences. It uses two feed-forward layers per transformer ``block,'' in addition to using ``delayed residuals'' which allows the model to choose which transformed representations to output.

GPT-2 is notable for being \textit{extremely deep}, with 1.5B parameters, 10x more than the original GPT architecture. This gives it more flexibility in learning tasks unsupervised from language modeling, especially when trained on a very large unlabeled corpus. 

\subsection{Multitask Finetuning}
BERT and GPT-2 both lack an explicit ``language model finetuning step,'' which gives ULMFiT an advantage where it learns to adapt to the writing style and linguistic features of the text used by its target task. Motivated by this, we propose to augment Transformer-based TL techniques with a language model finetuning step. 

Following recent works in multitask learning \cite{chronopoulou2019embarrassingly,liu2019multi}, we finetune the model to the writing style of the target task (via language modeling) \textit{at the same time} as we finetune the classifier, instead of setting it as a separate step. This produces two losses to be optimized \textit{together} during training, and ensures that no task (language modeling or classification) will be prioritized over the other. This concept has been proposed and explored to improve the performance of transfer learning in multiple language tasks .

We show that this method improves performance on both BERT and GPT-2, given that it learns to adapt to the idiosyncracies of its target task in a similar way that ULMFiT also does.

\section{Experimental Setup}
In this section, we describe how we preprocessed our datasets, pretrained our language models, finetuned them into fake news classifiers, and analyzed our benchmark results.

\subsection{Fake News Dataset Preprocessing}
To preprocess the dataset, we only perform tokenization, specifically ``Byte-Pair Encoding'' (BPE) \cite{bpe2018cherry}. 

BPE is a form of fixed-vocabulary \textit{subword tokenization} that considers \textit{subword units} as the most primitive form of entity (i.e. a \textit{token}) instead of canonical words (i.e. ``I am walking today'' $\rightarrow$ ``I am walk \#\#ing to \#\#day''). BPE is useful as it allows our model to represent out-of-vocabulary (OOV) words unlike standard tokenization. In addition, it helps language models in learning morphologically-rich languages as it now treats morphemes as the smallest token forms instead of whole words.

For training/finetuning the classifiers, we use a 70\%-30\% train-test split of the dataset. To produce the splits, we load the full dataset into Pandas, shuffle the dataset in-place using the \texttt{sample} function, setting \texttt{frac} to 1.0 and the random seed to 42.

When finetuning for Transformer-based techniques, we do not adjust or trim the vocabulary of the model to the vocabulary of the dataset. Instead, we keep the original vocabulary used by the model.

\subsection{Pretraining Corpora}

\begin{table*}[t!]
\centering
\begin{tabular}{lllll}
\hline
  Split & Documents & Tokens & Unique Tokens & Num. of Lines \\ \hline
  Training & 120,975 & 39,267,089 & 279,153 & 1,403,147 \\
  Validation & 25,919 & 8,356,898 & 164,159 & 304,006 \\
  Testing & 25,921 & 8,333,288 & 175,999 & 298,974 \\
  \hline
  OOV Tokens & 28,469 (0.1020\%) & & \\
  \hline
  
\end{tabular}
\caption{\label{corpusstats} Statistics for the WikiText-TL-39 Dataset.}
\end{table*}
To pretrain BERT and GPT-2 language models, as well as an AWD-LSTM language model for use in ULMFiT, a large unlabeled training corpora is needed. For this purpose, we construct a corpus of 172,815 articles from Tagalog Wikipedia\footnote{https://tl.wikipedia.org/wiki/Unang\_Pahina} which we call \textit{WikiText-TL-39} \cite{wiki2019cruz}. 

We form training-validation-test splits of 70\%-15\%-15\% from this corpora. To form the splits, we load the full dataset into Pandas, shuffle in place (like in the Fake News Filipino dataset) to form 70\%-30\% splits, then further splitting the 30\% split in half to form the validation and test splits. In computing for the dataset statistics, we do not consider tokens in the validation set as part of the training set vocabulary.

In constructing the dataset, the text was pre-split using the Moses Tokenizer \cite{koehn2007moses} such that it can be space-split for use in later tasks like in the original WikiText datasets \cite{merity2016pointer}. For use in pretraining, however, the dataset was preprocessed lightly similar to the FFN datatset, and was tokenized using Byte-Pair Encoding.

Corpus statistics for the pretraining corpora are shown on table \ref{corpusstats}. For more information on dataset construction and benchmarks, we refer the reader to \cite{wiki2019cruz}.

\subsection{Siamese Network Training}
We train a siamese recurrent neural network as our baseline. For each twin, we use 300 dimensions for the embedding layer and a hidden size of 512 for all hidden state vectors. 

To optimize the network, we use a regularized cross-entropy objective of the following form:
\begin{equation}
	\begin{split}
	\mathcal{L}(x_1, x_2) = \textnormal{y}(x_1, x_2)\log \textnormal{p}(x_1, x_2) + \\
 	(1 - \textnormal{y}(x_1, x_2))\log (1 - \textnormal{p}(x_1, x_2)) + \lambda |w|^2
 	\end{split}
\end{equation}
where y$(x_1, x_2)$ = 1 when $x_1$ and $x_2$ are from the same class and 0 otherwise. We use the Adam optimizer \cite{kingmaAdam2014} with an initial learning rate of 1e-4 to train the network for a maximum of 500 epochs.

\subsection{Transfer Pretraining}
We pretrain a cased BERT-Base model using our prepared unlabeled text corpora using Google's provided pretraining scripts\footnote{https://github.com/google-research/bert}. For the masked language model pretraining objective, we use a 0.15 probability of a word being masked. We also set the maximum number of masked language model predictions to 20, and a maximum sequence length of 512. For training, we use a learning rate of 1e-4 and a batch size of 256. We train the model for 1,000,000 steps with 10,000 steps of learning rate warmup for 157 hours on a Google Cloud Tensor processing Unit (TPU) v3-8. 

For GPT-2, we pretrain a GPT-2 Transformer model on our prepared text corpora using language modeling as its sole pretraining task, according to the specifications of \cite{gpt22019radford}. We use an embedding dimension of 410, a hidden dimension of 2100, and a maximum sequence length of 256. We use 10 attention heads per multihead attention block, with 16 blocks composing the encoder of the transformer. We use dropout on all linear layers to a probability of 0.1. We initialize all parameters to a standard deviation of 0.02. For training, we use a learning rate of 2.5e-4, and a batch size of 32, much smaller than BERT considering the large size of the model. We train the model for 200 epochs with 1,000 steps of learning rate warmup using the Adam optimizer. The model was pretrained for 178 hours on a machine with one NVIDIA Tesla V100 GPU.

For ULMFiT, we pretrain a 3-layer AWD-LSTM  model with an embedding size of 400 and a hidden size of 1150. We set the dropout values for the embedding, the RNN input, the hidden-to-hidden transition, and the RNN output to (0.1, 0.3, 0.3, 0.4) respectively. We use a weight dropout of 0.5 on the LSTM's recurrent weight matrices. The model was trained for 30 epochs with a learning rate of 1e-3, a batch size of 128, and a weight decay of 0.1. We use the Adam optimizer and use slanted triangular learning rate schedules \cite{ulmfit2018howard}. We train the model on a machine with one NVIDIA Tesla V100 GPU for a total of 11 hours.

For each pretraining scheme, we checkpoint models every epoch to preserve a copy of the weights such that we may restore them once the model starts overfitting. This is done as an extra regularization technique.

\subsection{Finetuning}
We finetune our models to the target fake news classification task using the pretrained weights with an appended classification layer or \textit{head}.

For BERT, we append a classification head composed of a single linear layer to the transformer model. We then finetune our BERT-Base model on the fake news classification task for 3 epochs, using a batch size of 32, and a learning rate of 2e-5.

For GPT-2, our classification head is first comprised of a layer normalization transform, followed by a linear layer. We finetune the pretrained GPT-2 transformer for 3 epochs, using a batch size of 32, and a learning rate of 3e-5.

For ULMFiT, we perform language model finetuning on the fake news dataset (appending no extra classification heads yet) for a total of 10 epochs, using a learning rate of 1e-2, a batch size of 80, and weight decay of 0.3. For the final ULMFiT finetuning stage, we append a compound classification head (linear $\rightarrow$ batch normalization $\rightarrow$ ReLU $\rightarrow$ linear $\rightarrow$ batch normalization). We then finetune for 5 epochs, gradually unfreezing layers from the last to the first until all layers are unfrozen on the fourth epoch. We use a learning rate of 1e-2 and set Adam's $\alpha$ and $\beta$ parameters to 0.8 and 0.7, respectively.

To show the efficacy of Multitask Finetuning, we augment BERT and GPT-2 to use this finetuning setup with their classification heads, outputting logits for both classification and language modeling at the output layer. We finetune both models to the target task for 3 epochs, using a batch size of 32, and a learning rate of 3e-5. For optimization, we use Adam with linear warmup steps of 10\% the number of steps, comprising 3 epochs.

\subsection{Generalizability Across Domains}
To study the generalizability of the model to different news domains, we test our models against test cases not found in the training dataset. We mainly focus on three domains: political news, opinion articles, and entertainment/gossip articles. Articles used for testing are sourced from the same websites that the dataset was taken from.

\section{Results and Discussion}

\subsection{Classification Results}

Our baseline model, the siamese recurrent network, achieved an accuracy of 77.42\% on the test set of the fake news classification task.

The transfer learning methods gave comparable scores. BERT finetuned to a final 87.47\% accuracy, a 10.05\% improvement over the siamese network's performance. GPT-2 finetuned to a final accuracy of 90.99\%, a 13.57\% improvement from the baseline performance. ULMFiT finetuning gave a final accuracy of 91.59\%, an improvement of 14.17\% over the baseline Siamese Network.  

We could see that TL techniques outperformed the siamese network baseline, which we hypothesize is due to the intact pretrained knowledge in the language models used to finetune the classifiers. The pretraining step aided the model in forming relationships between text, and thus, performed better at stylometric based tasks with little finetuning.

The model results are all summarized in Table \ref{accresults}.

\begin{table*}[t!]
\begin{center}
\begin{tabular}{|l|l|l|l|l|l|}
\hline Model & Val. Accuracy & Loss & Val. Loss & Pretraining Time & Finetuning Time \\ \hline
Siamese Networks & 77.42\% & 0.5601 & 0.5329 & N/A & 4m per epoch \\
BERT & 87.47\% & 0.4655 & 0.4419 & 66 hours & 2m per epoch \\ 
GPT-2 & 90.99\% & 0.2172 & 0.1826 & 78 hours & 4m per epoch \\
ULMFiT & 91.59\% & 0.3750 & 0.1972 & 11 hours & 2m per epoch \\
\hline
ULMFiT (no LM Finetuning) & 78.11\% & 0.5512 & 0.5409 & 11 hours & 2m per epoch \\
\hline
BERT + Multitasking & 91.20\% & 0.3155 & 0.3023 & 66 hours & 4m per epoch \\ 
GPT-2 + Multitasking & 96.28\% & 0.2609 & 0.2197 & 78 hours & 5m per epoch \\
\hline
\end{tabular}
\end{center}
\caption{\label{accresults} Consolidated experiment results. The first section shows finetuning results for base transfer learning methods and the baseline siamese network. The second section shows results for ULMFiT without Language Model Finetuning. The last section shows finetuning results for transformer methods augmented with multitasking heads. BERT and GPT-2 were finetuned for three epochs in all cases and ULMFiT was finetuned for 5  during classifier finetuning.}
\end{table*}

\subsection{Language Model Finetuning Significance}
One of the most surprising results is that BERT and GPT-2 performed worse than ULMFiT in the fake news classification task despite being deeper models capable of more complex relationships between data. 

We hypothesize that ULMFiT achieved better accuracy because of its additional language model finetuning step. We provide evidence for this assumption with an additional experiment that shows a decrease in performance when the language model finetuning step is removed, droppping ULMFiT's accuracy to 78.11\%, making it only perform marginally better than the baseline model. Results for this experiment are outlined in Table \ref{accresults}.


In this finetuning stage, the model is said to ``adapt to the idiosyncracies of the task it is solving'' \cite{ulmfit2018howard}. Given that our techniques rely on linguistic cues and features to make accurate predictions, having the model adapt to the writing style of an article will therefore improve performance. 

\subsection{Multitask-based Finetuning}
We used a multitask finetuning technique over the standard finetuning steps for BERT and GPT-2, motivated by the advantage that language model finetuning provides to ULMFiT, and found that it greatly improves the performance of our models. 

BERT achieved a final accuracy of 91.20\%, now marginally comparable to ULMFiT's full performance. GPT-2, on the other hand, finetuned to a final accuracy of 96.28\%, a full 4.69\% improvement over the performance of ULMFiT. This provides evidence towards our hypothesis that a language model finetuning step will allow transformer-based TL techniques to perform better, given their inherent advantage in modeling complexity over more shallow models such as the AWD-LSTM used by ULMFiT. Rersults for this experiment are outlined in Table \ref{accresults}.


\section{Ablations and Further Discussions}
In this section, we perform ablations to study the performance contributions of key mechanisms and architectural choices in the models used. We also look into the behavior of the classifier when presented with articles outside its training set, as well as articles that aren't inherently fake or real.

\subsection{Pretraining Effects}
An ablation on pretraining was done to establish evidence that pretraining before finetuning accounts for a significant boost in performance over the baseline model. Using non-pretrained models, we finetune for the fake news classification task using the same settings as in the prior experiments.

In Table \ref{ablations}, it can be seen that generative pretraining via language modeling does account for a considerable amount of performance, constituting 44.32\% of the overall performance (a boost of 42.67\% in accuracy) in the multitasking setup, and constituting 43.93\% of the overall performance (a boost of 39.97\%) in the standard finetuning setup.

This provides evidence that the pretraining step is necessary in achieving state-of-the-art performance. 

\begin{table*}[t!]
\begin{center}
\begin{tabular}{|l|l|l|l|l|l|}
\hline Finetuning & Pretrained? & Accuracy & Val. Loss & Acc. Inc. & \% of Perf. \\ \hline
Multitasking & No & 53.61\% & 0.7217 & - & - \\
 & Yes & 96.28\% & 0.2197 & +42.67\% & 44.32\% \\ \hline
Standard & No & 51.02\% & 0.7024 & - & - \\
 & Yes & 90.99\% & 0.1826 & +39.97\% & 43.93\% \\
\hline
\end{tabular}
\end{center}
\caption{\label{ablations} An ablation study on the effects of pretraining for multitasking-based and standard GPT-2 finetuning. Results show that pretraining greatly accounts for almost half of performance on both finetuning techniques. ``Acc. Inc.'' refers to the boost in performance contributed by the pretraining step. ``\% of Perf.'' refers to the percentage of the total performance that the pretraining step contributes.}
\end{table*}

\subsection{Attention Head Effects}
An ablation was performed to study the contribution of the multiheaded nature of the attention mechanisms to the final performance. 

\begin{table}[t!]
\begin{center}
\begin{tabular}{|l|l|l|l|}
\hline \# of Heads & Accuracy & Val. Loss & Effect \\ \hline
1 & 89.44\% & 0.2811 & -6.84\% \\
2 & 91.20\% & 0.2692 & -5.08\%  \\
4 & 93.85\% & 0.2481 & -2.43\% \\
8 & 96.02\% & 0.2257 & -0.26\% \\
10 & 96.28\% & 0.2197 &  \\
16 & 96.32\% & 0.2190 & +0.04 \\
\hline
\end{tabular}
\end{center}
\caption{\label{heads} An ablation study on the effect of multiple heads in the attention mechanisms. The results show that increasing the number of heads improves performance, though this plateaus at 10 attention heads. All ablations use the multitask-based finetuning method. ``Effect'' refers to the increase or decrease of accuracy as the heads are removed. Note that 10 heads is the default used throughout the study.}
\end{table}

For this experiment, we performed several pretraining-finetuning setups with varied numbers of attention heads using the multitask-based finetuning scheme. Using a pretrained GPT-2 model, attention heads were masked with zero-tensors to downsample the number of positions the model could attend to at one time. 

As shown in Table \ref{heads}, reducing the number of attention heads severely decreases multitasking performance. Using only one attention head, thereby attending to only one context position at once, degrades the performance to less than the performance of 10 heads using the standard finetuning scheme. This shows that more attention heads, thereby attending to multiple different contexts at once, is important to boosting performance to state-of-the-art results.

While increasing the number of attention heads improves performance, keeping on adding extra heads will not result to an equivalent boost as the performance plateaus after a number of heads. 

\begin{figure}[htb]
\centering
    \includegraphics[width=8cm]{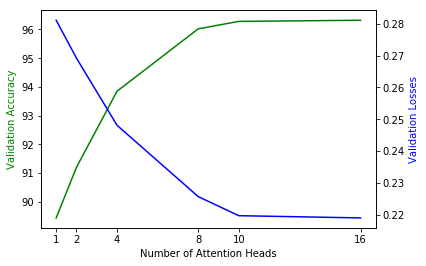}
    \caption{Ablation showing accuracy and loss curves with respect to attention heads.}
    \label{fig:ablationheads}
\end{figure}

As shown in Figure \ref{fig:ablationheads}, the performance boost of the model plateaus after 10 attention heads, which was the default used in the study. While the performance of 16 heads is greater than 10, it is only a marginal improvement, and does not justify the added costs to training with more attention heads.

\subsection{Generalizability Across Domains}
When testing on three different types of articles (Political News, Opinion, Entertainment/Gossip), we find that writing style is a prominent indicator for fake articles, supporting previous findings regarding writing style in fake news detection \cite{stylometric2018potthast}.

When given actual labeled fake and real news articles, our best performing model was able to classify them properly. Here are some excerpts taken outside the training dataset, one fake article (Figure \ref{sample2}) and one real article (Figure \ref{sample1}):

\begin{figure}[!htb]
\noindent
\framebox{\parbox{\dimexpr\linewidth-2\fboxsep-2\fboxrule}{
May bagong paandar ang mga anti-DU30, at ito ay pag-po-post ng mga black ang white videos ng mga peronalidad sa mundo ng showbiz at sining kung saan pinahahapyawan nila ang Duterte Administration. Ngunit sa serye ng mga videos na kanilang inilathala sa kanilang Facebook page, ang video ni Director Joel Lamangan ang pinagpiyestahan ng husto ng mga netizens. \\

Ayon kasi kay lamangan, walang respeto daw ang kasalukuyang gobyerno at iboto daw ang mga taga-oposisyon para mapatunayan ang pagiging tunay na Pilipino. \\


\noindent\rule{5cm}{0.4pt}

\textit{Translation: The Anti-Duterte camp has a brand new new tactic, which involves posting black and white videos of famous personalities in showbusiness and the arts where they criticize the Duterte Administration. However, the among the videos that they put out on their Facebook page, the video of Director Joel Lamangan drew the most flak and was feasted upon by netizens.} \\

\textit{According to Lamangan, the current government has no respect and the people should vote for the opposition in order to show that they are true Filipinos.} \\


\textbf{Classifier Label: Fake} \\
\textbf{Gold label: Fake}
}}
\caption{\label{sample2} Example of a fake article from a known fake news source. The classifier labels this as fake.}
\end{figure}

\begin{figure}[!htb]
\noindent
\framebox{\parbox{\dimexpr\linewidth-2\fboxsep-2\fboxrule}{
Inaresto ng anti-scalawag and intelligence units ng Philippine National Police (PNP) ang isang anti-drug operative ng Pasay City Police Station ngayong araw, matapos na isangkot sa extortion at kidnapping. \\ \\
Dinampot si Police Corporal Anwar Nasser, nakatalaga sa Station Drug Enforcement Team (SDET) ng Pasay police, ng mga tauhan ng PNP-Counter Intelligence Task Force (CITF) at Intelligence Group (IG). \\

Samantala, tatlo sa kanyang mga kasabwat ang nakatakas. Kinilala ang mga ito na sina Police Lieutenant Ronaldo Frades, hepe ng Pasay SDEU; Patrolman Anthony Fernandez; at Sergeant Rigor Octaviano, na pawang miyembro ng Pasay SDET. \\


\noindent\rule{5cm}{0.4pt}

\textit{Translation: The anti-scalawag and intelligence units of the Philippine National Police (PNP) arrested an anti-drug operative of the Pasay City Police Station today after being involved i in extortion and kidnapping.} \\

\textit{Police Corporal Anwar Nasser, part of the Station Drug Enforcement Team (SDET) of the Pasay police, was arrested by members of the PNP-Counter Intelligence Task Force (CITF) and the Intelligence Group (IG). } \\

\textit{Three of his accomplices, however, were able to escape. They are identified as Police Lieutenant Ronaldo Frades, chief of the SDEU; Patrolman Anthony Fernandez; and Sergeant Rigor Octaviano, who was also a member of the Pasay SDET.} \\


\textbf{Classifier Label: Real} \\
\textbf{Gold label: Real}
}}

\caption{\label{sample1} Example of a news article from a mainstream news source. The classifier labels this as real.}
\end{figure}

Following \cite{FactCheck}, fake news tends to be written in an ``insinuative'' or ``clickbait-y'' style, while real news is written to be more declarative of facts. These are styles that the above examples follow and that the classifier picks up on.

To further add evidence that the classifier picks up on style, we present two more excerpts, this time of a gossip article (Figure \ref{sample3}) and an opinion article (Figure \ref{sample4}):

\begin{figure}[!htb]
\noindent
\framebox{\parbox{\dimexpr\linewidth-2\fboxsep-2\fboxrule}{
Puro mapapaklang komento ang ipinakakain ngayon sa isang pamosong female personality ng mismong mga tagahanga niya. Pinasukan ng kawalan ng utang na loob at pangtatraydor pa nga ang kanilang emosyon. Sila raw ang dahilan kung bakit natupad ang pangarap ng babaeng personalidad na sumikat, mas nadagdagan daw ang naiipon niyang pera dahil sa kanyang mga tagasubaybay, pero ano ang ginawa sa kanila? \\ \\
Kuwento ng aming source, ``Iyakan sila nang iyakan! Sana raw, e, pinatay na lang sila ng idolo nila para isahan na lang ang sakit na naramdaman nila!''
``Tinatawag siyang walang utang na loob, nagkakandagutom daw sila para sa pagsuporta sa idol nila, wala naman silang hinihinging kapalit, pero ang napala nila?'' \\

\noindent\rule{5cm}{0.4pt}

\textit{Translation: Bitter comments are being served to a certain famous female personality by none other than her own fans. They were filled with traitorous emotions and a lack of indebtedness.They claim that they were the reason why the female personality's dreams came true, as well as why she's earning money. But what was done to them in the first place?} \\

\textit{According to our source, ``They were all crying! They wished that their idol just killed them on the spot so they don't have to feel any more pain! They say she has no indebtedness, they starve just to support their idol, they weren't asking for anything in return, but what happened to them?''} \\

\textbf{Classifier Label: Fake} \\
\textbf{Gold label: --}
}}

\caption{\label{sample3} Example of a gossip (blind item) article from a mainstream tabloid. Notice that the classifier labels it as fake, despite not being inherently fake news.}
\end{figure}

\begin{figure}[!htb]
\noindent
\framebox{\parbox{\dimexpr\linewidth-2\fboxsep-2\fboxrule}{
NANININDIGAN ang Malacanang sa desisyon na payagan ang Department of Interior and Local Government (DILG) na ilabas ang pangalan ng mga kandidato na sangkot sa illegal drugs. Para sa Malacanang, makatutulong sa mga botante ang paglalabas ng ``narcolist'' para makapamili ng iluluklok sa puwesto. Handa naman daw ang Malacanang sa mga magsasampa ng kaso kaugnay sa paglalabas ng ``narcolist''. Karapatan daw ito ng sinuman. Mabuti nga iyon para malinis ang pangalan. \\ 

Maraming bumabatikos sa idea ng DILG at Philippine Drug Enforcement Agency (PDEA) na ilabas ang mga pangalan ng kandidatong sangkot sa illegal drugs. Labag umano ito sa Konstitusyon. Ang nararapat daw gawin ng DILG o ng Malacanang ay sampahan ng kaso ang mga kandidatong ayon sa kanila ay sangkot sa illegal drugs. Maaaring magamit daw ito ng mga kalaban sa pulitika at paano kung hindi naman totoo ang paratang. Nasira na raw ang reputasyon ng kandidato at maaaring hindi na ito manalo. Sino pa ang boboto sa taong dinurog na ang pagkatao. Dapat daw ay mag-isip muna nang malalim bago ihayag ang mga pangalan ng kandidatong sangkot sa droga. \\

\noindent\rule{5cm}{0.4pt}

\textit{Translation: Malacanang was adamant in their decision to allow the Department of Interior and Local Government (DILG) to release the names of election candidates who are allegedly involved in illegal drugs. For Malacanang, this would help the people in choosing who to vote. Malacanang said they were ready for anyone who sues them in relation to the release of this ``narcolist'' and that this was the right of anyone. This would be good in order to clean their names.} \\

\textit{People are criticizing this idea of the DILG and the Philippine Drug Enforcement Agency (PDEA), however. According to common consensus, this was against the constitution. The right thing to do instead is to sue the people on the list directly, as the list can be used against political enemies, tarnishing their names even if the link is not proven. Such opposition may lose. After all, who would vote for someone with an unclean reputation? The people call on the government to think twice before releasing this list.} \\

\textbf{Classifier Label: Real} \\
\textbf{Gold label: --}
}}

\caption{\label{sample4} Example of an opinion article from a mainstream news organization. Notice that the classifier tags it as real despite being opinionated.}
\end{figure}

While the two articles are not necessarily real or fake, their writing styles cause the classifier to label them as such. Gossip articles (at least in Philippine media) are naturally written to catch attention with a more ``clickbait-y'' and insinuative style of writing, akin to how fake news articles are written. This causes the gossip article to be classified as fake. In the same way, an opinion or editorial article is more commonly written with facts in a declarative manner, before the author or editor's own opinions are added along with supporting details. This causes the article to be labeled as real.

\section{Conclusion}
Fake news detection remains to be an important task to tackle given the speed that fake news travels across social media and other mediums throughout the internet. While methods to detect fake news exist, they all require a large amount of news samples and supporting data (including, but is not limited to: fact checks, source verifications, and stance information). For most, if not all, low-resource languages, such data and resources are not available. This requires researchers dealing with low-resource languages to employ different techniques to produce robust classifiers with the little data available to them.

In this work, we alleviate scarcity in the low-resource Filipino language by introducing the Fake News Filipino (FFN) dataset, the first full benchmark dataset for fake news detection in Filipino. We then benchmark transfer learning methods and show that robust fake news classifiers can be produced using such techniques, augmenting transformer-based techniques to adapt to writing style via multitask learning. 

Furthermore, we perform ablations and showed that each technique and architecture choice in the models contribute towards the robust performance of the models. We also studied the behavior of our best performing model when given articles beyond its training dataset. We show that the model adequately captures and classified via style, classifying gossip and opinion articles as fake and real respectively despite not being inherently fake or real articles.

While we present robust results in this work, this does not mean that low-resource fake news detection has been solved. Aside from increasing the size of the fake news dataset for future research, we recommend the following for further work:

On the domain level, we recommend investigating classifier behavior on other types of news articles including satire, news with misleading headlines, and incompletely-written news, types of articles that we were not able to study due to the lack of expertly-annotated and curated data of such kind in the Philippines. 

On the task level, we suggest augmenting basic content-based fake news detection with other tasks such as content-based stance verification \cite{confortiStance2018} and headline-stance vs content stance detection \cite{stance2017bourgonje}. While we posit that learning these tasks would make more robust fake news detectors, it includes the caveat that more data must be collected and curated, the type of which may not be present in the Philippines, or in other countries with low-resource languages.

Lastly, on the model level, we suggest investigating smaller models with comparable performance to the models benchmarked in this work, as the models we used cannot be easily deployed for consumer-level applications that could allow a wide audience of users to quickly verify fake and real news articles.


\section*{Acknowledgments}
The authors would like to acknowledge the efforts of VeraFiles and the National Union of Journalists in the Philippines (NUJP) for their work covering and combating the spread of fake news. 

We are partially supported by Google's Tensoflow Research Cloud (TFRC) program. Access to the TPU units provided by the program allowed the BERT models in this paper, as well as the countless experiments that brought it to fruition, possible.

\section{Bibliographical References}
\bibliographystyle{lrec}
\bibliography{lrec2020W-xample}


\end{document}